\documentclass[conference]{IEEEtran}

\usepackage{amsmath,amssymb, graphicx}
\usepackage[ruled,linesnumbered]{algorithm2e}
\usepackage{algorithmic}
\usepackage{paralist} 
\usepackage{color}
\usepackage{multirow}
\usepackage{rotating}
\usepackage{hyperref}
\usepackage[mathscr]{eucal} 
\usepackage{amsbsy} 
\usepackage{booktabs}
\usepackage{xcolor}
\usepackage{tikz}
\usetikzlibrary{snakes}
\usepackage{multirow}

\usepackage[skins]{tcolorbox}
\usepackage{array}
\newcolumntype{P}[1]{>{\centering\arraybackslash}p{#1}}
\usepackage{wrapfig}
\usepackage{hhline}
\usepackage{hyperref}

\usepackage{subcaption} 
\usepackage{graphicx}
\usepackage{epstopdf}
\usepackage{balance}
\usepackage{tablefootnote}
\usepackage{empheq} 
\usepackage{moresize}
\usepackage{enumitem}
\setlist{nolistsep}

\usepackage[font=small,labelfont=bf,skip=0pt]{caption}

\DeclareCaptionType{copyrightbox}
\usepackage{url}

\newcounter{ALC@tempcntr}
\newcommand{\hide}[1]{}

\newcommand{\ben}{\begin{enumerate*}}
\newcommand{\een}{\end{enumerate*}}
\newcommand{\bit}{\begin{itemize*}}
\newcommand{\eit}{\end{itemize*}}

\def\x{{\mathbf x}}

\begin{document}

\author{\IEEEauthorblockN{Gisel Bastidas Guacho}
\IEEEauthorblockA{
UC Riverside\\
gbast001@ucr.edu}
\and
\IEEEauthorblockN{Sara Abdali}
\IEEEauthorblockA{
UC Riverside\\
sabda005@ucr.edu}
\and
\IEEEauthorblockN{Neil Shah}
\IEEEauthorblockA{Snap Inc.\\
nshah@snap.com}
\and
\IEEEauthorblockN{Evangelos E. Papalexakis}
\IEEEauthorblockA{
UC Riverside\\
epapalex@cs.ucr.edu}
}


\title{Semi-supervised Content-based Detection of Misinformation via Tensor Embeddings}
%

\maketitle
%
%
 \begin{abstract}

Fake news may be intentionally created to promote economic, political and social interests, and can lead to negative impacts on humans beliefs and decisions. Hence, detection of fake news is an emerging problem that has become extremely prevalent during the last few years. Most existing works on this topic focus on manual feature extraction and supervised classification models leveraging a large number of labeled (fake or real) articles. In contrast, we focus on content-based detection of fake news articles, while assuming that we have a \emph{small} amount of labels, made available by manual fact-checkers or automated sources.  We argue this is a more realistic setting in the presence of massive amounts of content, most of which cannot be easily fact-checked. To that end, we represent collections of news articles as multi-dimensional tensors, leverage tensor decomposition to derive concise article embeddings that capture spatial/contextual information about each news article, and use those embeddings to create an article-by-article graph on which we propagate limited labels. Results on three real-world datasets show that our method performs on par or better than existing models that are fully supervised, in that we achieve better detection accuracy using fewer labels. In particular, our proposed method achieves 75.43\% of accuracy using only 30\% of labels of a public dataset while an SVM-based classifier achieved 67.43\%. Furthermore, our method achieves 70.92\% of accuracy in a large dataset using only 2\% of labels.
\end{abstract}
\section{Introduction}

\label{sec:intro}
Misinformation on the web is a problem that has been greatly amplified by the use of social media, and the problem of fake news in particular has become ever more prevalent during the last years. Social media is a common platform for consuming and sharing news, due to its ease-of-use in diffusing content and promoting exposure/discussion. In fact, two-thirds of Americans reported getting some of their news from social media in 2017 \footnote{http://www.journalism.org/2017/09/07/news-use-across-social-media-platforms-2017/}. Even though social media has become a news source for its advantages, it is especially vulnerable to the propagation of fake news mostly coming from unverified publishers and crowd-based content creators because there is practically no control over the information that is shared. The well-documented spread of misinformation on Twitter during events such as the 2010 earthquake in Chile \cite{Mendoza:2010:TUC:1964858.1964869}, Hurricane Sandy in 2012 \cite{Gupta:2013:FSC:2487788.2488033}, the Boston Marathon blasts in 2013 \cite{Gupta:Boston13} and US Presidential Elections on Facebook in 2016\cite{Silverman:2016} are all such examples.
Since misinformation is intentionally created for malicious purposes such as obtain economic and political benefits or deceiving the public \cite{Shu:2017}, it can clearly lead to negative user experience by either influencing their beliefs and impacting their decisions for the worse.

Several approaches in recent literature have been proposed to automatically detect misinformation using supervised classification models. In \cite{Castillo:2011:ICT:1963405.1963500}, authors assess the credibility of a tweet based on Twitter features using a decision-tree based algorithm. Other works extract manually crafted features from news content such as the number of nouns, length of the article, fraction of positive/negative words, and more in order to discriminate fake news articles \cite{hardalov_koychev_nakov_2016,Rubin:2016,Horne:2017}.  Rubin et al. \cite{Rubin:2016} use five linguistic features as input to  an SVM-based classifier. Similarly, Horne and Adali \cite{Horne:2017} apply SVM classification on content-based features from three categories: stylistic, complexity and psychological features. In addition to these works, several others  proposed propagation-based models for evaluating news credibility \cite{Gupta:Han:2012,NewsVerif,Jin:2014}. Nonetheless, they initialized credibility values for the entire network using a supervised classifier.  However, the reality is that such labels are often very limited and sparse.  Fact-checking websites such as Snopes.com, PolitiFact.com, and FactCheck.org can be used to assess claims, but these websites require domain experts to assign credibility values to claims and are therefore, limited by human capacity.  Moreover, fact-checking is a time-consuming process, often requiring surveying multiple articles and sources, evaluating reputation and likelihood of the claims before coming to a decision. 

\begin{figure}[t!]
  \includegraphics[width=\linewidth]{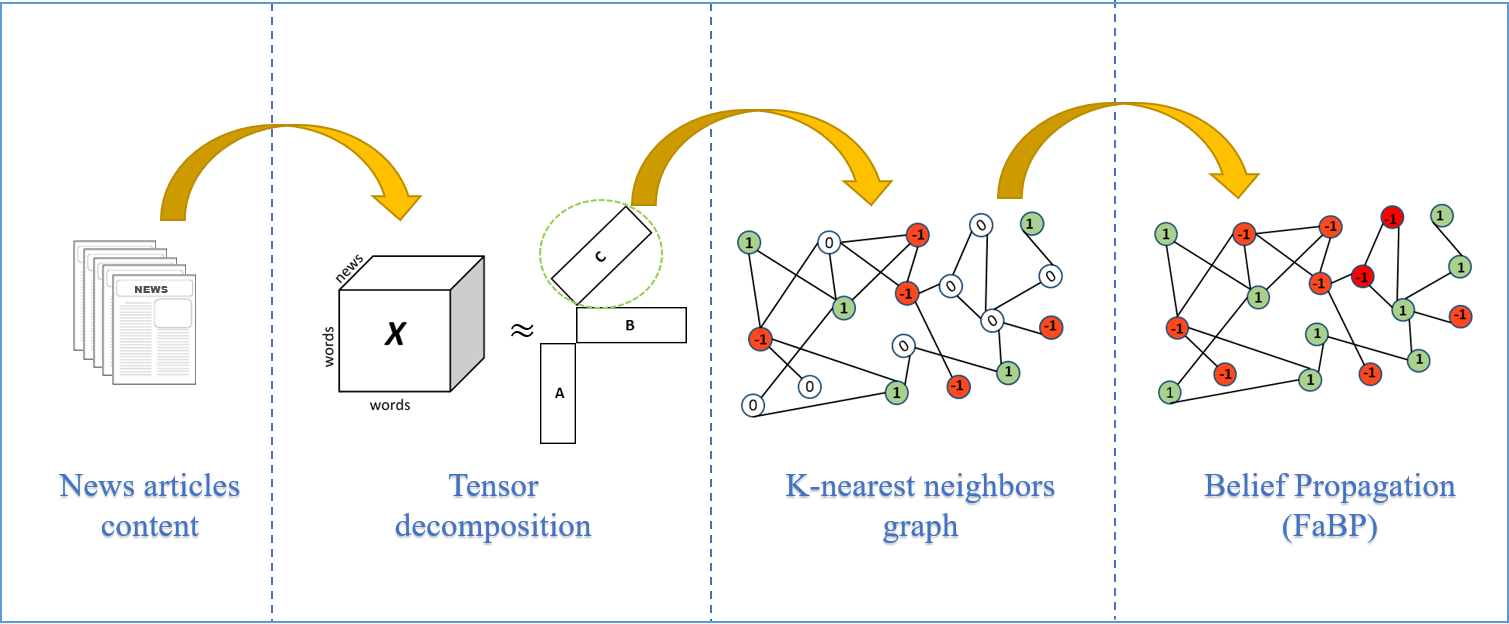}
  \centering
  \caption{Our proposed method discerns real from misinformative news articles via leveraging tensor representation and semi-supervised learning in graphs.}
  \label{fig:model}
\end{figure}

In this paper, we propose a new \emph{semi-supervised} approach for fake news article detection based on news content, which requires \emph{limited} labels to detect misinformation with high accuracy.  On a high level, our approach exploits tensor representation and decomposition of news articles, careful construction of a $k$-nearest neighbor graph, and propagation of limited labeled article information to conduct inference on a larger set.

Our main contributions are:
 \begin{itemize}
\item We leverage tensor-based article embeddings, which are shown to produce concise representations of articles with respect to their spatial context, in order to derive a graph representation of news articles. 
\item We formulate fake news detection as a semi-supervised method that propagates known labels on a graph to determine unknown labels.
\item We collect a large dataset of misinformation and real news articles publicly shared on social media.
 \item We evaluate our method on real datasets. Experiments on two previously used datasets demonstrate that our method outperforms prior works since it requires a fewer number of known labels and achieves comparable performance. In fact, our method attains $75.43\%$ accuracy using only $30\%$ of labels of the first public dataset and  $67.38\%$ accuracy using only $10\%$ of labels of the second public dataset. 
\end{itemize}

The remainder of this paper is organized as follows. In Section \ref{sec:problem}, we describe preliminaries and definitions used throughout this paper. Section \ref{sec:problemdef} defines the problem we address. In Section \ref{sec:method}, we introduce our proposed method in detail. In Section \ref{sec:experiments}, we describe the datasets, evaluation metrics and discuss the experimental results and performance of our approach. Section \ref{sec:sensitivity} involves sensitivity analysis, where we discuss the impact of type and length of news articles on the performance of the proposed method. 
Section \ref{sec:related} discusses related work, and Section \ref{sec:conclusions} concludes.

\section{Preliminaries and Notations}
\label{sec:problem}
In this section, we provide preliminary
definitions for technical concepts that we use throughout this paper.
\subsection{CP/PARAFAC Tensor Decomposition}
A tensor is a multidimensional array where its dimensions are referred as modes.
The most widely used tensor decomposition is Canonical Polyadic (CP) or PARAFAC decomposition \cite{harshman1970fpp}. CP/PARAFAC factorizes a tensor into a sum of rank-one tensors. For instance, a three-mode tensor is decomposed into a sum of outer products of three vectors: $$ \boldsymbol{\mathcal{X}} \approx \Sigma_{r=1}^{R} \mathbf{a}_r \circ \mathbf{b}_r \circ \mathbf{c}_r$$
where $\mathbf{a}_r \in \mathbb{R}^{I}$, $\mathbf{b}_r \in \mathbb{R}^{J}$, $\mathbf{c}_r \in \mathbb{R}^{K}$ and the outer product is given by  $$( \mathbf{a}_r , \mathbf{b}_r , \mathbf{c}_r)(i,j,k)= \mathbf{a}_r(i) \mathbf{b}_r(j) \mathbf{c}_r(k) \text{ for all } i,j,k \text{ \cite{Papalexakis:2016}}$$
The factor matrices are defined as
\begin{equation*} 
\begin{split}
\mathbf{A} & = [\mathbf{a}_1 ~ \mathbf{a}_2 \ldots \mathbf{a}_R]  \\
\mathbf{B} & = [\mathbf{b}_1 ~ \mathbf{b}_2 \ldots \mathbf{b}_R] \\
\mathbf{C} & = [\mathbf{c}_1 ~ \mathbf{c}_2 \ldots \mathbf{c}_R]
\end{split}
\end{equation*}
 where $\mathbf{A} \in \mathbb{R}^{I\times R}$, $\mathbf{B} \in \mathbb{R}^{J  \times R}$, and $\mathbf{C} \in \mathbb{R}^{K \times R}$ denote the factor matrices and $R$ is the rank of the decomposition, that is, the number of columns in the factor matrices. In this paper, we use CP/PARAFAC decomposition to decompose a three-mode tensor, though the decomposition supports higher modes as well.

\subsection{$k$-nearest-neighbor graph} 
A $k$-nearest-neighbor, or $k$-NN graph is one in which node $\boldsymbol p$ and node $\boldsymbol q$ are connected by an edge if node $\boldsymbol p$ is in $\boldsymbol q$'s $k$-nearest-neighborhood
or node $\boldsymbol q$ is in $\boldsymbol p$'s k-nearest-neighborhood. $k$ is a hyperparameter that determines the density of the graph -- thus, a graph constructed with small $k$ may be sparse or poorly connected.

The $k$-nearest-neighbors of a point in $n$-dimensional space are defined using a ``closeness'' relation where proximity is often defined in terms of a distance metric \cite{Han:2011:DMC:1972541} such as Euclidean $\ell_2$ distance. Thus, given a set of points $\mathcal{P}$ in $n$-dimensional space, a $k$-NN graph on $\mathcal{P}$ can be constructed by computing the $\ell_2$ distance between each pair of points and connecting each point with the $k$ most proximal ones.

The $\ell_2$ distance $d$ between two points $\boldsymbol p$ and $\boldsymbol q$ in $n$-dimensional space is defined as:
$$d(p,q)=\sqrt[]{\sum_{i=1}^n(q_i-p_i)^2} $$




\subsection{Fast Belief Propagation (FaBP)}
FaBP \cite{KoutraKKCPF11} is a fast and linearized guilt-by-association method, which improves upon the basic idea of belief propagation (BP) over a graph.  In our case, a belief is the label of a news article. Hence, we use FaBP as a means to propagate label likelihood over a graph, given set of known labels.  The operative intuition behind FaBP and other such guilt-by-association methods is that nodes which are ``close'' are likely to have similar labels or belief values.    

The FaBP algorithm solves the following linear system:  $$[\mathbf{I} + a \mathbf{D} - c'\mathbf{A}]b_h = \phi_h$$
where $\phi_h$ and $b_h$ denote prior and final beliefs, respectively. $\mathbf{A}$ denotes the $n \times n$ adjacency matrix of an underlying graph of $n$ nodes, $\mathbf{I}$ denotes the $n \times n$ identity matrix, and $\mathbf{D}$ is a $n \times n$ diagonal matrix of degrees where $\mathbf{D}_{ii}= \sum_j \mathbf{A}_{ij}$ and $\mathbf{D}_{ij}=0$ for $i \neq j$. Finally, we define $a=\frac{4h_h^2}{1-4h_h^2}$ and $c'= \frac{2h_h}{(1-4h_h^2)}$ where $h_h$ denotes the homophily factor between nodes (i.e. their ``coupling strength'' or association).  More specifically, higher homophily means that close nodes tend to have more similar labels. The coefficient values are set as above for convergence reasons; we refer the interested reader to \cite{KoutraKKCPF11} for further discussion.  

\section{Problem Definition} 
\label{sec:problemdef}
We consider a misinformative, or fake, news article as one that is \textit{``intentionally and verifiably false,''} following the definition used in \cite{Shu:2017}.  With this definition in mind, we aim to discern fake news articles from real ones based on their content. Henceforth, by ``content,'' we refer to the text of the article. We reserve the investigation of other types of content (such as image and video) for future work. 

Let $\mathcal{N}=\{n_1,n_2,n_3,...,n_M\}$ be a collection of news articles of size $M$ where each news article is a set of words and $\mathcal{D} = \{ w_1, w_2,w_3, ..., w_I \}$ be a dictionary of words of size $I$. Assuming that labels of some news articles are available. Let $l \in \{-1,0,1\}$ denote a vector containing the partially known labels, such that entries of $1$ represent real articles, $-1$ represents fake articles and $0$ denotes an unknown status.  Using the aforementioned notations, we formally define the problem as follows:
\tcbset{colback=yellow!5!white,colframe=yellow!30!white}
\begin{tcolorbox}[width=\linewidth]
\textbf{Given} a collection of news articles $\mathcal{N}$ and a label vector $l$ with entries for labeled real/fake and unknown articles, we aim to \textbf{predict} the class labels of the unknown articles.
\end{tcolorbox}

We address the problem as a binary classification problem; hence, a news article is classified either fake or real.

\section{Proposed Method}
\label{sec:method}
 In this section, we introduce a content-based method for semi-supervised fake news article detection. This method consists of three steps: representing/decomposing content as a tensor, constructing a $k$-NN graph of proximal embeddings, and propagating beliefs using FaBP. Figure \ref{fig:model} summarizes our proposed method. Below, we describe the individual steps in further detail.
 
\subsection*{Step 1: Tensor Decomposition.} 
 Given a collection of news articles $\mathcal{N}=\{n_1,n_2,n_3,...,n_M\}$ of size $M$, where each news article in $\mathcal{N}$ is a vector that contains the words within the news article,
we build similar tensor-based article embeddings as proposed in \cite{Hosseinimotlagh2017UnsupervisedCI}.  Specifically, we propose the use of two different types of tensor construction methods, which we will further compare and evaluate in Sections \ref{sec:experiments} and \ref{sec:sensitivity}.
 
\subsubsection{Frequency-based tensor} 
We build a three-mode tensor $\boldsymbol{\mathcal{X}} \in \mathbb{R}^{I\times I\times M}$ $(words, words, news)$ where for each news article, we create a co-occurrence matrix with non-zeros indicating $(word_1, word_2)$ appearances within a window parameter of $w$ (usually 5-10) words \footnote{We experimented with small values of that window and results were qualitatively similar.}. More specifically,  $\boldsymbol{\mathcal{X}}(i,j,k)$  contains the number of times that the $i^{th}$ and $j^{th}$ words appear within the predefined window in the news article $k$.
\subsubsection{Binary-based tensor} We additionally build a tensor $\boldsymbol{\mathcal{Y}} \in \mathbb{R}^{I\times I\times M}$ $(words, words, news)$ where all co-occurrence entries are boolean and indicate indicates whether the $i^{th}$ and $j^{th}$ words appeared within the predefined window at least once.

We then use CP/PARAFAC tensor decomposition \cite{harshman1970fpp} to factorize the tensors.  As \cite{Hosseinimotlagh2017UnsupervisedCI} demonstrates, such tensor-based article embeddings captures spatial/contextual nuances of different types of news articles and result in homogeneous article groups. After decomposing the tensor, we obtain the factor matrices $ \mathbf{A}, \mathbf{B}, \mathbf{C}$ whose columns correspond to different latent topics, clustering news articles and words in the latent topic space. More specifically, each row of $\mathbf{C}$ is the representation of the corresponding article in the resulting embedding space.
 
\subsection*{Step 2: $k$-NN graph of news articles.}
The tensor embedding we computed in Step 1 provides a compact and discriminative representation of news articles into a concise set of latent topics. Using this embedding, we construct a graphical representation of news articles. In particular, we use the factor matrix $\mathbf{C}$ in order to construct a $k$-NN graph $G$ of news articles. As we mentioned before each column in $\mathbf{C}$ is the representation of the corresponding news article in the latent topic space; thus, by constructing a $k$-NN graph on $\mathbf{C}$, we can find similar articles in that space. To this end, we consider each row in $\mathbf{C} \in \mathbb{R}^{M\times R}$ as a point in $R$-dimensional space. We then compute $\ell_2$ distance among news and find the $k$-closest points for each point in $\mathbf{C}$. Since the number of news articles can be extremely large in practice, we can leverage well-known $kd$-tree based optimizations \cite{pelleg1999accelerating} in order to more efficiently find to $k$-nearest-neighbors for each article.

Each node in $G$ represents a news article and each edge encodes that two articles are similar in the embedding space.  Since we use a distance metric, if article $n_1$ is in the $k$-nearest-neighbors of $n_2$, it does not mean that $n_2$ is in the $k$-nearest-neighbors of $n_1$, since $n_1$ may have other neighbors which are closer than $n_2$. However, in this step, we only leverage the distance as a means to measure similarity between news articles, without much concern for the actual order of proximity.  Thus, we enforce symmetry in the neighborhood relations, that is, if $n_1$ is a $k$-nearest-neighbor of news $n_2$, the opposite should also hold.  The resultant graph $G$ is an undirected, symmetric graph where each node is connected to at least $k$ nodes. The graph can be compactly represented as an $M \times M$ adjacency matrix.

\subsection*{Step 3: Belief Propagation.} 
 Using the graphical representation of the news articles above, and considering that for a small set of those news articles we have ground truth labels, our problem becomes an instance of semi-supervised learning over graphs. We use a belief propagation algorithm which assumes homophily, because news articles that are connected in the $k$-NN graph are likely to be of the same type due to the construction method of the tensor embeddings; moreover, \cite{Hosseinimotlagh2017UnsupervisedCI} demonstrates that such embeddings produces fairly homogeneous article groups. More specifically, we use the fast and linearized FaBP variant proposed in \cite{KoutraKKCPF11}. The algorithm is demonstrated to be insensitive to the magnitude of the known labels; thus, we consider that FaBP can achieve good learning performance only using a small number of known labels. Moreover, FaBP can be used for large-scale graphs since it is linear on the number of edges in $G$.  Hence, we classify fake news articles in a semi-supervised fashion.

\section{Experimental Evaluation}
\label{sec:experiments}
In this section, we evaluate the performance of our method on several real datasets. We implemented our method in MATLAB using the Tensor Toolbox \cite{matlab} and leveraged existing MATLAB FaBP implementation \cite{KoutraKKCPF11}.
\begin{table}[htbp]
\caption{Dataset specifics.}
\centering
 \begin{tabular}{llll} 
 \toprule
\textbf{Datasets} & \textbf{\# fake news} & \textbf{\# real news} & \textbf{\# total} \\ \midrule
Dataset1 (Political) & 75 & 75  & 150  \\ \addlinespace[3px]
Dataset2 (Bulgarian)  & 69 & 68 & 137\\ \addlinespace[3px]
 Our dataset& 31,739  &  31,739& 63,478 \\
 \bottomrule
\end{tabular}
\label{table:tab2}
\end{table}
\subsection{Dataset description} 
In order to evaluate the performance of our proposed method, we used data from three datasets: two public datasets of hundreds of articles, and our own, collected dataset of tens of thousands of articles, as shown in Table \ref{table:tab2}.

\subsubsection{Public datasets} The two public datasets were used in previous studies. Specifically, \emph{Dataset1} consists of 150 political news articles, balanced to have 75 articles of each class, and was provided by \cite{Horne:2017}. \emph{Dataset2} contains 68 real and 69 fake news articles, and was provided by \cite{hardalov_koychev_nakov_2016}.

\subsubsection{Our dataset} In constructing our dataset, we collected news article URLs from Twitter tweets during a 3-month period from June-August 2017. These URLs were filtered based on website domain. We then crawled those URLs to get the news article content. To that end, we created a crawler in Python that extracted news content using the web API boilerpipe\footnote{http://boilerpipe-web.appspot.com/} and the Python library Newspaper3k \footnote{http://newspaper.readthedocs.io/en/latest/}.
There were few cases where these tools did not extract the news content. For those cases, we also used Diffbot \footnote{https://www.diffbot.com/dev/docs/article/} which is
another API to extract article text from web pages.
All real news articles were featured on 367 domains obtained from Alexa \footnote{https://www.alexa.com/}, and fake news articles belong to 367 domains from the BSDetector (browser extension to identify fake news sites) domain list \cite{bsdetector}. Table \ref{table:tab3} describes the
BSDetector-specified domain categories. In this study, we considered news from those domain categories as fake news in order to later perform the binary classification task. All in all, the dataset consists of 31,739 fake news and 409,076 real news articles. Since this dataset is highly imbalanced, we randomly down-sampled the real class to get a balanced dataset. 


The content of each news article for the three datasets was preprocessed using tokenization and stemming. Besides, stopwords and punctuations were removed.
 
 \begin{table}[htbp]
\centering
\caption{Domain categories collected from BSDetector \cite{bsdetector}, as indicated in our dataset.}
 \begin{tabular}{p{1.5cm} p{5.5cm}} 
 \toprule
  \textbf{Category} &  \textbf{Description}  \\ 
 \midrule
 \emph{Bias} & Sources that traffic in political propaganda and gross distortions of fact.  \\ \addlinespace[3px]
 
  \emph{Clickbait} & Sources that are aimed at generating online advertising revenue and rely on sensationalist headlines or eye-catching pictures.\\ \addlinespace[3px]
  
 \emph{Conspiracy} & Sources that are well-known promoters of kooky conspiracy theories.\\ \addlinespace[3px]
 
\emph{Fake} & Sources that fabricate stories out of whole cloth with the intent of pranking the public. \\ \addlinespace[3px]

\emph{Hate} & Sources that actively promote racism, misogyny, homophobia, and other forms of discrimination. \\ \addlinespace[3px]

\emph{Junk Science} & Sources that promote pseudoscience, metaphysics, naturalistic fallacies, and other scientifically dubious claims.\\ \addlinespace[3px]

\emph{Rumor} & Sources that traffic in rumors, innuendo, and unverified claims.  \\ \addlinespace[3px]

\emph{Satire} &Sources that provide humorous commentary on current events in the form of fake news. \\
 \bottomrule
\end{tabular}
\label{table:tab3}
\end{table} 

\begin{figure}[htbp]
  \includegraphics[width=0.90\linewidth]{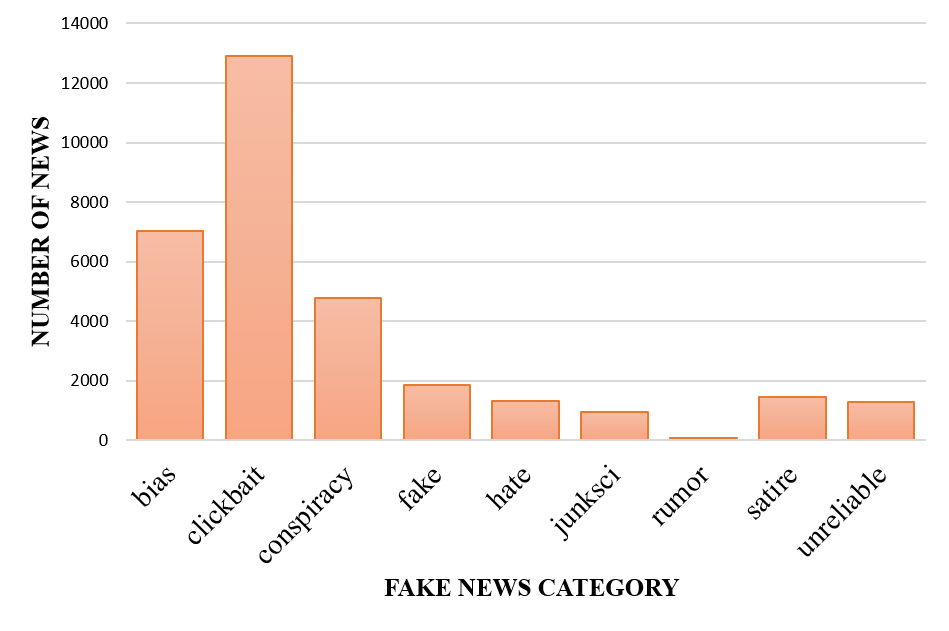}
  \centering
  \caption{Distribution of misinformation per domain category in our collected dataset.}
  \small
  \label{fig:distFn}
\end{figure}
\subsection{Evaluation Metrics}
Since we consider the fake article detection as a binary classification problem, we evaluated our method in terms of the following commonly used metrics:
{\footnotesize$$Accuracy=\frac{|TP|+|TN|}{|TP|+|TN|+|FP|+|FN|},$$} which is the percentage of correctly classified articles.
{\footnotesize$$Precision =\frac{|TP|}{|TP|+|FP|},$$}
which is the percentage of articles predicted as fake out of all articles predicted as fake.
{\footnotesize$$Recall=\frac{|TP|}{|TP|+|FN|},$$} which is the percentage of all fake articles that are correctly predicted as fake. 
{\footnotesize$$F1\_score=2 \cdot \frac{Precision \cdot Recall}{Precision+Recall},$$} which is the harmonic mean of precision and recall, and indicates a combined measure of performance.

$TP$ denotes true positives (correctly predicted fake news articles), $FP$ denotes false positives (real news articles which were predicted as fake), $TN$ denotes true negatives (i.e. correctly predicted real news articles), and $FN$ denotes false negatives (i.e. fake news articles which were predicted as real).

\subsection{Evaluation}

\begin{figure}[htbp]
  \includegraphics[width=\linewidth]{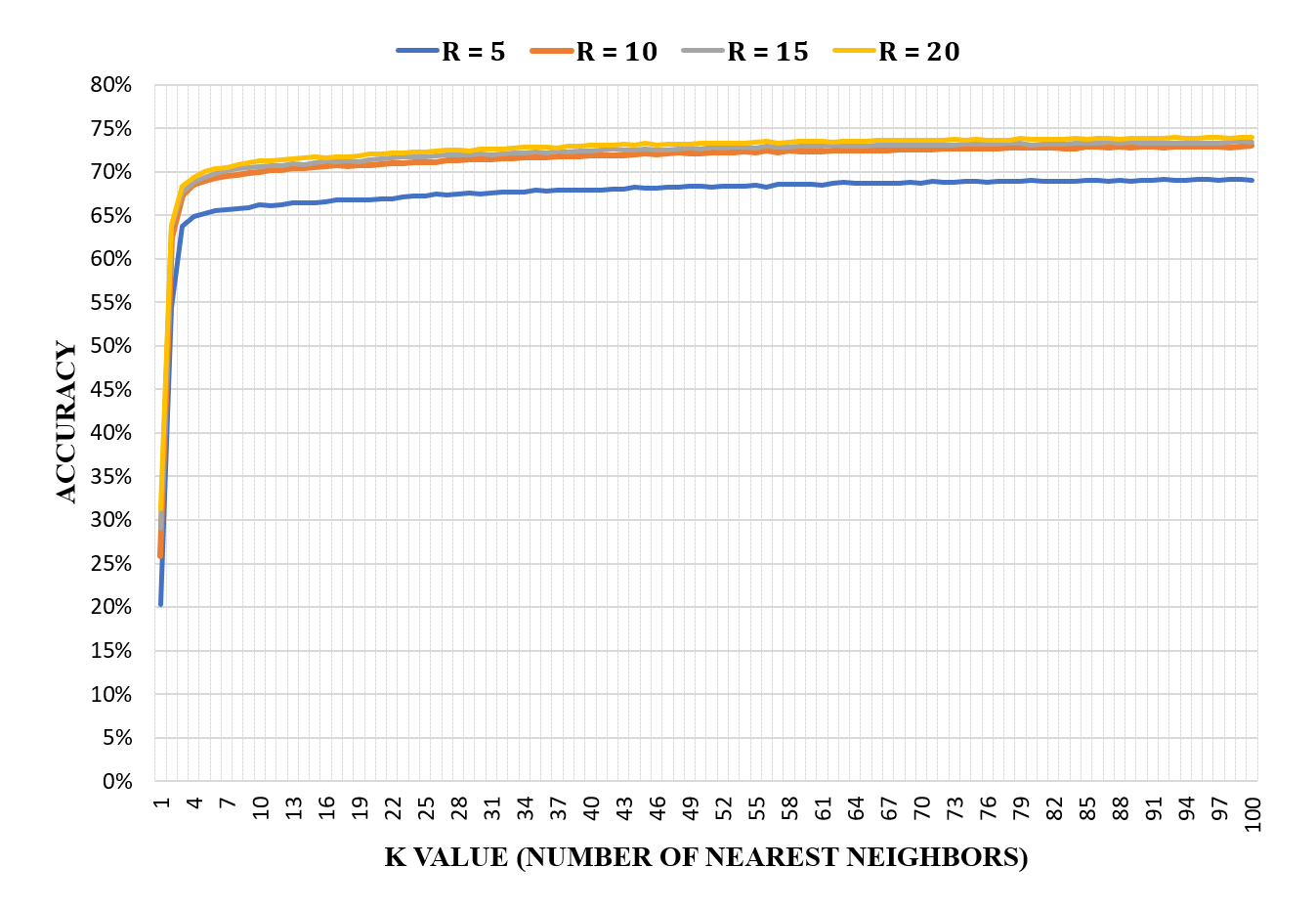}
  \centering
  \caption{Performance using different parameter settings for decomposition rank ($R$) and number of nearest neighbors ($k$).}
  \label{fig:parameters}
\end{figure}

In order to find the best-performing parameters for our method,
we run an iterative process using cross-validation where we evaluated different settings with respect to $R$ (i.e. decomposition rank) and  $k$ (i.e. the number of nearest neighbors, controlling the density of the $k$-NN graph $G$).
We considered values of $R$ from 1 to 20, since decomposition rank is often set to be low for time and space reasons in practice \cite{shah2014spotting}. Likewise, we tested $k$ with values from 1 to 100, trading off greater bias for less variance with increasing $k$.
We found that the best accuracy is obtained when both parameters $R$ and $k$ are set to be 10. We find that for values of $k$ and $R$ greater than 10, performance is qualitatively similar as shown in Figure \ref{fig:parameters}, and thus we fix the parameters as such in evaluation. Notice that using a small $k$ value (for example, 1 or 2), the accuracy is relatively poor; this is because building a $k$-NN graph with small $k$ results in a highly sparse graph which offers limited propagation capacity. In all experiments, we tested accuracy over the test set of all articles whose labels were ``unknown'' or unspecified in the propagation step. 

\begin{figure}[htbp]
  \includegraphics[width=0.90\linewidth]{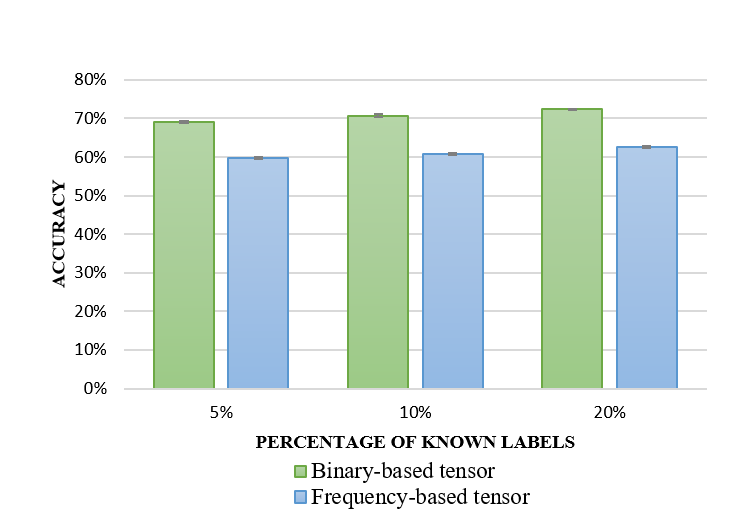}
  \centering
  \caption{Detection performance using different tensor representations.}
  \label{fig:acc2}
\end{figure}

As mentioned in Section \ref{sec:method}, we consider two tensor construction methods: \emph{frequency-based} and \emph{binary-based}. Figure \ref{fig:acc2} shows the comparative performance of our method using both of these. We find that binary-based tensors perform better than frequency-based ones in the detection task. Thus, we used binary-based representations in evaluation. We discuss intuition for why binary-based tensors result in improved performance in Section \ref{sec:sensitivity}.

\begin{table*}[htbp]
\centering
\caption{Performance of the proposed method using our dataset with different percentages of labeled news.}
 \begin{tabular}[lcr]{lllll}
 \toprule
\textbf{\%Labels} & \textbf{Accuracy}& \textbf{Precision} & \textbf{Recall} & \textbf{F1} \\
 \midrule
5\% & 69.12 $\pm$ 0.0026 & 69.09 $\pm$ 0.0043  & 69.24 $\pm$ 0.0090  & 69.16 $\pm$  0.0036 \\ \addlinespace[3px]
10\% & 70.76 $\pm$ 0.0027 &  70.59 $\pm$ 0.0029  &  71.13 $\pm$ 0.0101  &  70.85 $\pm$ 0.0043 \\ \addlinespace[3px]
20\% & 72.39 $\pm$ 0.0013 & 71.95 $\pm$ 0.0017 & 73.32 $\pm$ 0.0043& 72.63 $\pm$ 0.0017\\ \addlinespace[3px]
30\% & 73.44 $\pm$ 0.0008 & 73.13 $\pm$ 0.0028  & 74.14 $\pm$ 0.0034  & 73.63 $\pm$ 0.0007\\ 
 \bottomrule
\end{tabular}
\label{table:tab1}
\end{table*}

\begin{figure}[htbp]
  \includegraphics[width=0.88\linewidth]{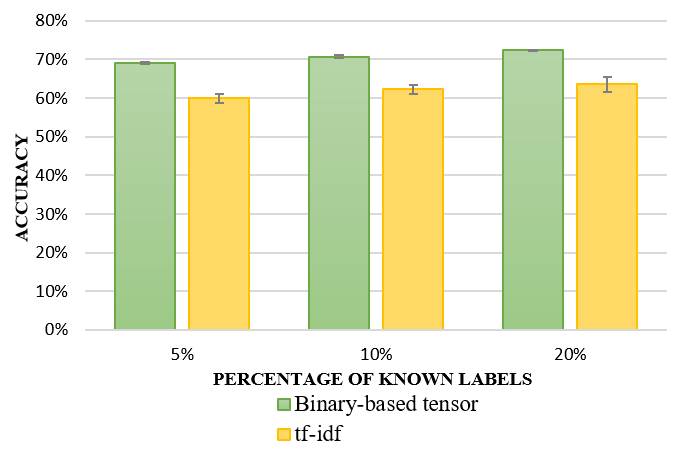}
  \centering
  \caption{Performance our method using tensor-based article embedding compared to Belief propagation using a graph built form $tf$-$idf$ matrix.}
  \label{fig:acc3}
\end{figure}

We evaluated our method with different percentages $p$ of known labels. Table \ref{table:tab1} shows the performance of our method using $p \in \{5\%, 10\%, 20\%, 30\% \}$ of labeled news articles from our dataset. Our results demonstrate that we can achieve an accuracy of 70.76\% only using 10\% of labeled articles. Additionally, to evaluate the quality of tensor embeddings over traditional vectorial representations, we compared performance between our approach and a variant in which between we constructed a $k$-NN graph built from the term frequency inverse-document-frequency ($tf$-$idf$) representations.  Figure \ref{fig:acc3} shows that our method with tensor embeddings consistently attains better accuracy than the alternative over varying known label percentages. This empirically suggests that binary-based tensor representations can better captures spatial/contextual nuances of news articles over vectorial representations. 

\begin{figure}[htbp]
  \includegraphics[width=0.90\linewidth]{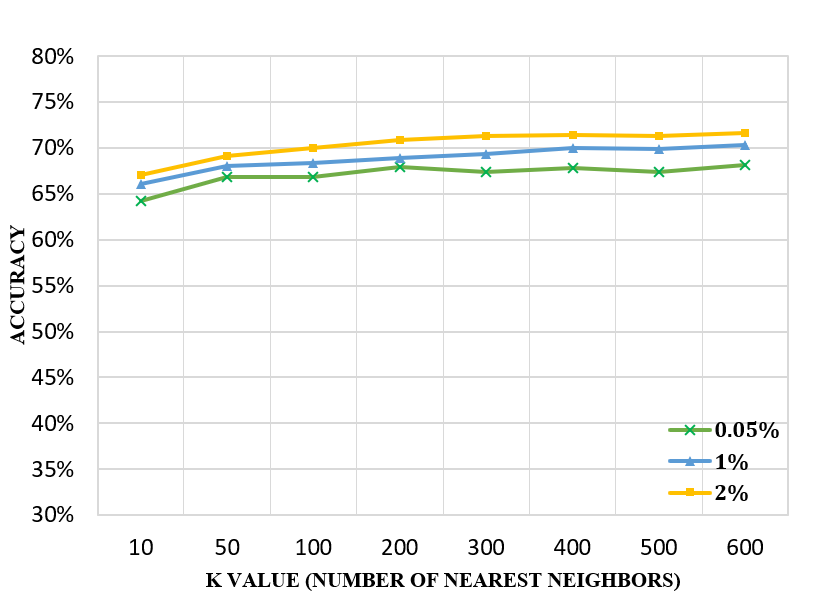}
  \centering
  \caption{Performance using extremely sparse ($<$5\%) labeled articles and varying number of nearest neighbors.}
  \label{fig:prl5}
\end{figure}

To evaluate the performance of our approach using extremely sparse known labels, we also evaluated our method using percentages of known labels $<$5\% and varying the number of nearest neighbors.  Figure \ref{fig:prl5} shows that we can achieve an accuracy of 70.92\% using 2\% of known labels when the number of nearest neighbors is set to be 200.  In fact, the performance of our approach degrades fairly gracefully with even smaller proportions of known labels.

\begin{figure}[h]
  \includegraphics[width=0.90\linewidth]{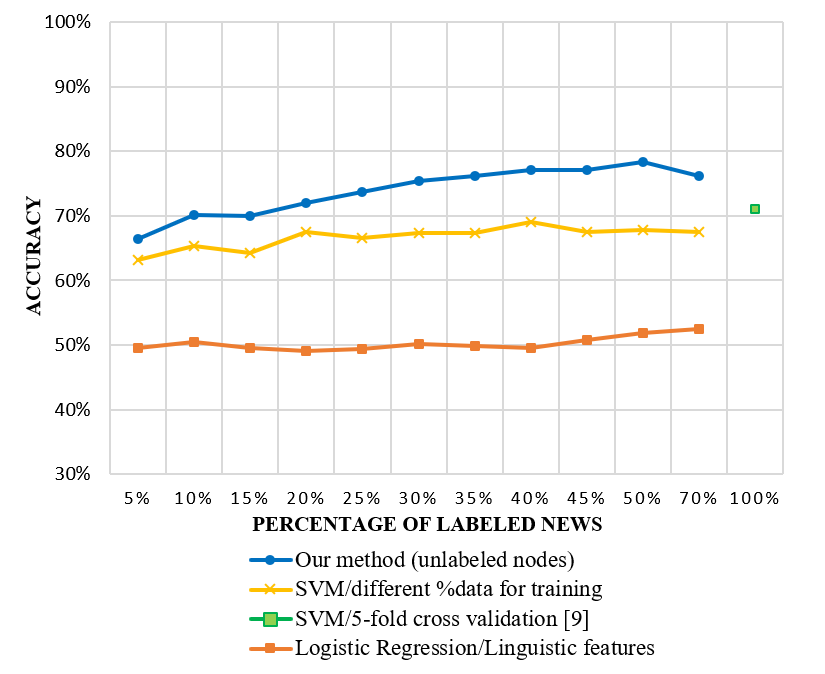}
  \centering
  \caption{Performance using \emph{Dataset1} provided by Horne et al. \cite{Horne:2017} }
  \label{fig:ds1}
\end{figure}

In addition, we evaluated our model using \emph{Dataset1}  and \emph{Dataset2}.  We compare the accuracy achieved by our method to the accuracy achieved by the following approaches:
\begin{itemize}
\item \emph{SVM on content-based features} as proposed in  \cite{Horne:2017}.  To this extent, we replicated the feature extraction from news content and used SVM in order to show the performance using different percentages of data to train the model. 
\item \emph{Logistic regression on content-based features} proposed by \cite{hardalov_koychev_nakov_2016}. We used their publicly available implementation. In particular, we run their method with linguistic ($n$-gram) feature extraction using different percentages of data to train their model. 
\end{itemize}
Figure \ref{fig:ds1} shows the results for   \emph{Dataset1}. Our approach demonstrates improved accuracy even with fewer labels  -- specifically, we achieved $75.43\%$ accuracy using only the $30\%$ of news labels while SVM($30\%$/$70\% $ train/test), SVM(5-fold cross-validation), and logistic regression ($30\%$/$70\%$ train/test) attained $67.43\%$, $71\%$ and $50.09\%$ of accuracy, respectively. The accuracy achieved by SVM(5-fold cross-validation) was reported by Horne et al. in \cite{Horne:2017}.

For \emph{Dataset2}, we run logistic regression and SVM, using $10\%$/$90\%$ train/test split.  These approaches achieved an accuracy of $59.84\%$ and $64.79\%$, respectively, compared to the $67.38\%$ accuracy achieved by our approach, using the same percentage of labeled articles. 



We note that our method is able to achieve this performance only having a small number of labeled news articles due to the semi-supervised nature of our approach; leveraging the $k$-NN graph and propagating beliefs allows us to exploit similarity between even unlabeled news articles, which supervised classification systems cannot leverage.

\section{Sensitivity Analysis}
\label{sec:sensitivity}

In this section, we discuss how news article length and category/domain affect the performance of our method. In particular, we study performance impact when considering varying news article lengths or categories, for both frequency-based and binary-based tensor embeddings.  To this end, we created sub-sampled datasets from our original dataset meeting the following conditions:
 \begin{itemize}
 \item News articles have similar content length and are selected across news categories
 \item  News articles vary in length and belong to the same news category
 \item News articles have similar content length and belong to the same news category
 \end{itemize}
 \begin{figure}[h]
  \includegraphics[width=0.90\linewidth]{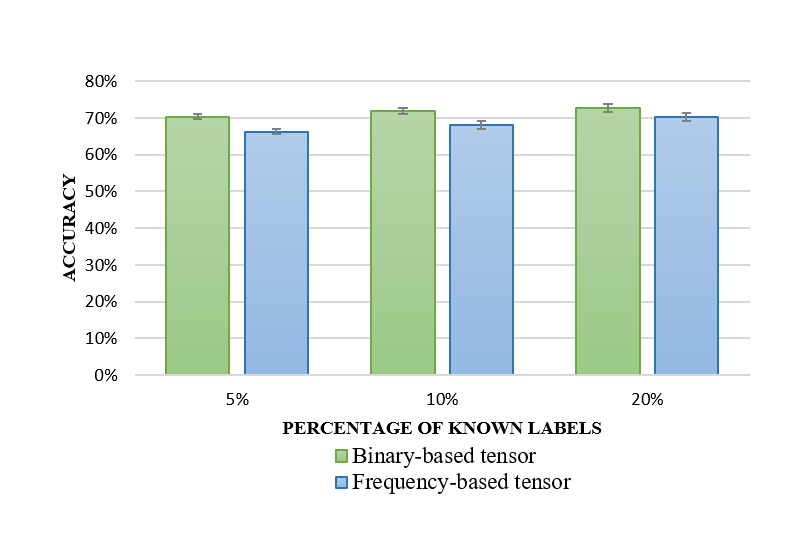}
  \centering
  \caption{Performance using binary vs. frequency-based tensor embeddings on news articles of all types with similar content length.}
  \label{fig:stacc2}
\end{figure}
\begin{figure*}[tb]
    \centering
    \begin{subfigure}[]{\columnwidth}
       \includegraphics[width=0.85\textwidth ]{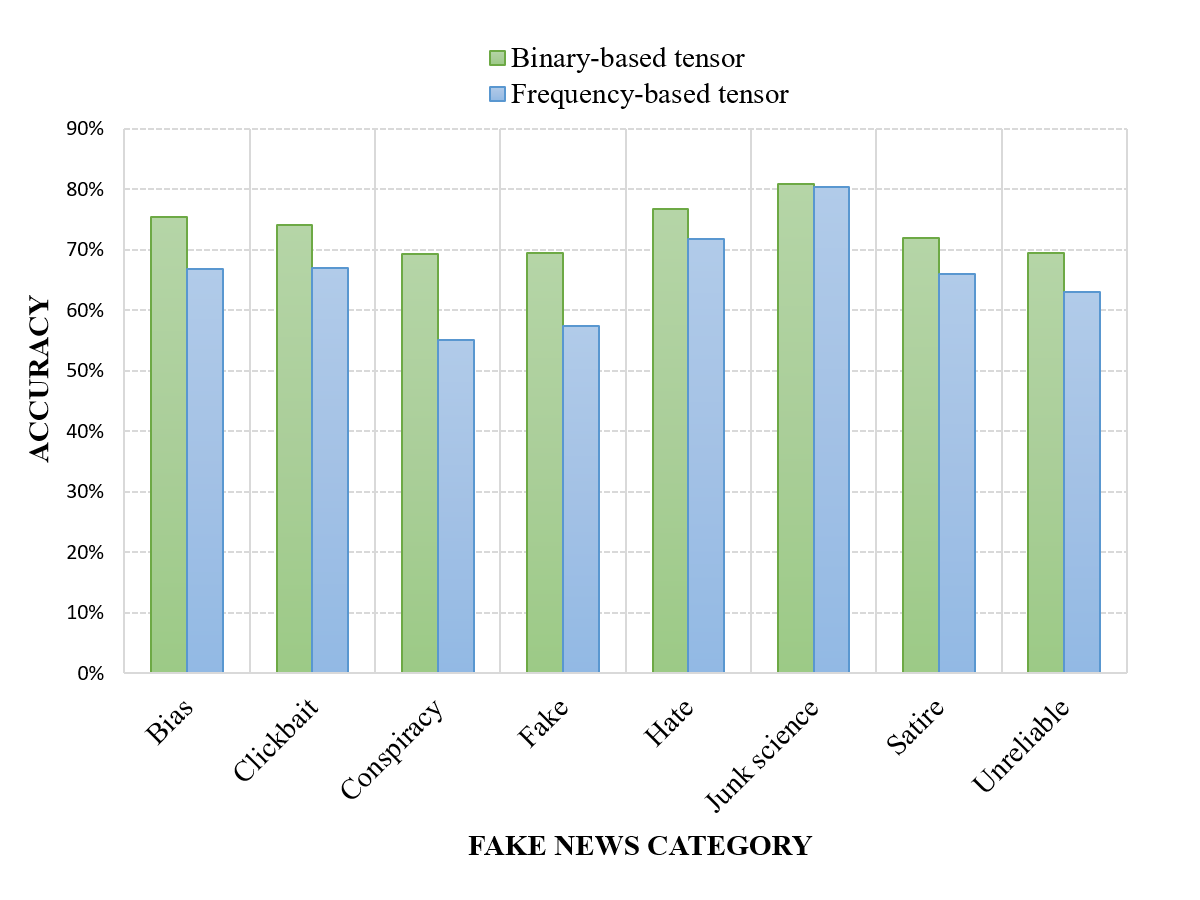}
      \caption{}
        \label{fig:sp1}
    \end{subfigure}%
~~~~~~~~~~~~~~
    \begin{subfigure}[]{\columnwidth}
        \includegraphics[width=0.85\textwidth] {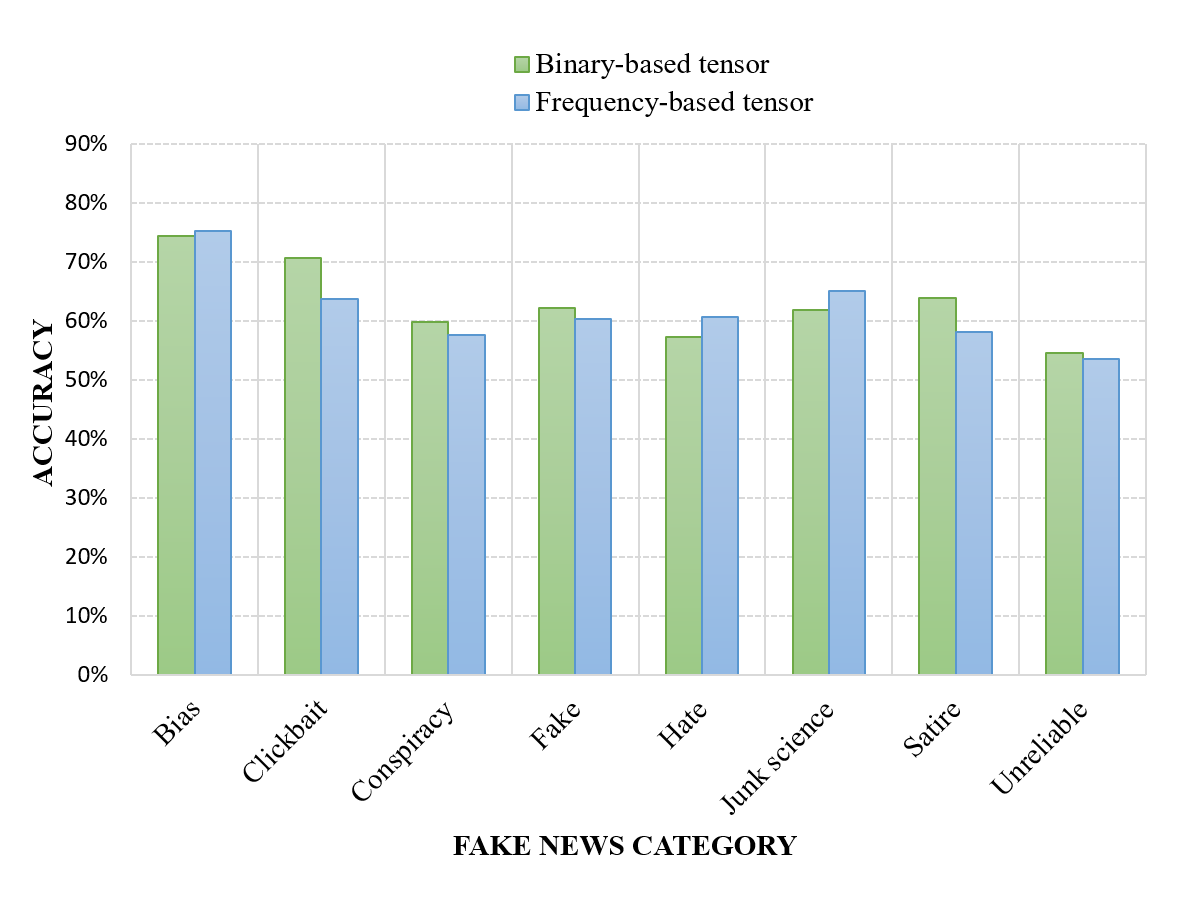}
        \caption{}
        \label{fig:sp2}
    \end{subfigure}

    \caption{Performance using binary vs. frequency-based tensor embeddings on category-partitioned news articles (a) varying in length, and (b) with similar length.}\label{fig:sp}
\end{figure*}

We first evaluated our method using a dataset where news articles have similar content length and fake articles are selected across news categories (Table \ref{tab:tabType} shows summary statistics for article length across each fake news category). 

Figure \ref{fig:stacc2} shows the accuracy achieved by our method using both binary-based and frequency-based tensor embeddings.  The results suggest that performance is not especially sensitive to news category when length is standardized.

In addition, we evaluated our method using 8 sample datasets, one for each misinformative news type: \emph{bias}, \emph{clickbait}, \emph{conspiracy}, \emph{fake}, \emph{hate}, \emph{junk science}, \emph{satire}, and \emph{unreliable}. Each dataset was balanced, containing the same number of fake and real articles. Note that in these sample datasets, news article length varies (see Table \ref{tab:tabType}).

In Figure \ref{fig:sp1} we can observe the accuracy that our method achieved for each fake news type for both binary-based and frequency-based embeddings.  These results show that the binary-based representation noticeably outperforms the frequency-based in all categories. We then performed a comparable experiment, but instead only selected news articles that had almost the same length per category. Figure \ref{fig:sp2} shows that the results of experiment are less clear -- in fact, the results using both representations are quantitatively quite similar.  These findings indicate that the news articles length greatly affects how well the different co-occurrence tensor embeddings perform.  More specifically, we can conclude that binary-based tensors indicating  boolean co-occurrence between words better captures spatial/contextual nuances of news articles that vary in length. On the same note, detection performance is relatively comparable for embedding types when considering articles of a fixed or almost-fixed length.
\begin{table}[tb]
\centering
\caption{Dataset statistics per fake news category}
\begin{tabular}{llll} 
\toprule
\multirow{2}{*}{\textbf{Dataset}} & \multicolumn{3}{c}{\textbf{Article Length}} 
\\ \cmidrule[0.3px]{2-4}
   & \textbf{ Minimum} & \textbf{Mean} &  \textbf{Maximum} \\  
 \midrule
   \emph{Bias} & 18 & 363 &  5,903 \\ \addlinespace[3px]
   \emph{Clickbait} & 18 & 355 & 10,955 \\ \addlinespace[3px]
   \emph{Conspiracy} & 19 & 422 & 10,716 \\ \addlinespace[3px]
   \emph{Fake} & 20 & 378 & 8,803 \\ \addlinespace[3px]
   \emph{Hate} & 20 & 315 & 5,390 \\ \addlinespace[3px]
   \emph{Junk Science} & 22 & 364 & 5,390 \\ \addlinespace[3px]
   \emph{Satire} & 18 & 307 & 8,913\\ \addlinespace[3px]
   \emph{Unreliable} & 20 & 360& 5,268 \\
\bottomrule
\end{tabular}
\label{tab:tabType}
\end{table}
\section{Related Work}
\label{sec:related}
Below, we discuss related work in detecting misinformation.

\subsection{Supervised models} Several works that aim to detect misinformation are focused on supervised learning models based on extracted features from news content, user or contextual information.  \cite{Gupta:2012} proposed a ranking model based on SVM and Pseudo-Relevance Feedback to assess the credibility of information in tweets using content and source based features such as the number of unique characters, hashtags, and friend and follower counts. In \cite{hardalov_koychev_nakov_2016}, the authors proposed a logistic regression classifier using linguistic ($n$-gram), credibility (punctuation, pronoun use, capitalization) and semantic features generated from the news content.  \cite{Horne:2017} used SVM on content-based features that are categorized into stylistic, complexity and psychological features in order to classify real, fake and satirical news. In \cite{Qazvinian:2011}, the authors propose detecting rumors by building na\"{i}ve-Bayes classifiers on content, network and microblog-specific features.  \cite{Ma:2016} and \cite{DBLP:journals/corr/RuchanskySL17} leverage temporal structure by using recurrent neural network (RNN) based models to represent text and user characteristics. In \cite{Ma:2015}, the authors propose a Dynamic Series-Time Structure (DSTS) model for detecting rumors by capturing the social context of an event from content, user and propagation-based features. 
\vspace{2mm}

In contrary to the aforementioned works, we use news content in order to construct a tensor of co-occurrence between words for each news article. This tensor-based representation captures similarity in the context of the spatial relations among words between news articles.

\subsection{Propagation models} On the other hand, there are previous works that proposed propagation-based models for evaluating news credibility. \cite{Gupta:Han:2012} proposed a PageRank-like credibility propagation method on multi-typed network of events, tweets and users. 
\cite{NewsVerif} proposed constructing a credibility network for news verification based on conflicting viewpoints between tweets (i.e. based on positive and negative arguments about the news). In \cite{Jin:2014}, the authors proposed a hierarchical propagation model on a three-layer credibility network that consists of event, sub-event and message layers. These layers form a hierarchical structure that models their relations. These works all require initial credibility values, and obtain them by using the output of a supervised classifier. 
\vspace{2mm}

Unlike these works, we focus on a semi-supervised method that leverages tensor decomposition and belief propagation, and works well with very few labels.

\section{Conclusions}
\label{sec:conclusions}
In this paper, we propose a semi-supervised content-based method for detecting misinformative news articles. Our method leverages tensor-based article embeddings in order to construct a $k$-nearest neighbor graph of news articles which captures similarity between them in a latent, embedding space. We then use a guilt-by-association propagation algorithm to diffuse known article labels over the graph. We evaluate our method using our own dataset of over 63K articles, and two public datasets.  Experiments on these three real-world datasets demonstrate that our model is able to distinguish fake from real news only using a small number of labeled articles, compared to state-of-the-art content-based approaches which achieve similar quality while assuming fully supervised models. More specifically, our method achieves $75.43\%$ accuracy using only $30\%$ of labels of the first public dataset and  $67.38\%$ accuracy using only $10\%$ of labels of the second public dataset. Additionally, our method attains $70.92\%$ of accuracy using only $2\%$ of labels of our dataset. 



\hide{
\section{Acknowledgements}
{\scriptsize
Research was supported by the National Science Foundation Grant No. XXXXXX. Any opinions, findings, and conclusions or recommendations expressed in this material are those of the author(s) and do not necessarily reflect the views of the funding parties.
}
}

\balance
\bibliographystyle{IEEEtran}
\bibliography{BIB/vagelis_refs.bib}

\end{document}